\documentclass[runningheads]{llncs}
\usepackage[T1]{fontenc}
\usepackage{graphicx}
\usepackage{tcolorbox}
\usepackage{booktabs} 
\usepackage{multirow}
\usepackage{algorithm}
\usepackage{wrapfig}
\usepackage[noend]{algpseudocode}
\usepackage{lipsum}
\usepackage{float}
\usepackage{graphicx}
\usepackage{amsmath}
\usepackage[hyphens]{url}
\newtcolorbox{mybox}[1][]{
    title=#1,
    fonttitle=\small,
    fontupper=\small,
    left=2mm,
    right=2mm,
    top=1mm,
    bottom=0mm,
}

\usepackage{tikz}
\makeatletter
\newcommand{\DrawLine}{%
  \begin{tikzpicture}
  \path[use as bounding box] (0,0) -- (\linewidth,0);
  \draw[color=black,dashed,dash phase=2pt]
        (0-\kvtcb@leftlower-\kvtcb@boxsep,0)--
        (\linewidth+\kvtcb@rightlower+\kvtcb@boxsep,0);
  \end{tikzpicture}%
  }
\makeatother
\begin{document}
\title{Enhancing Retrieval Augmented Generation with Hierarchical Text Segmentation Chunking}

\author{{Hai-Toan Nguyen\thanks{Corresponding author}}, Tien-Dat Nguyen, \and
Viet-Ha Nguyen}
%
%
\institute{Institute for Artificial Intelligence, VNU University of Engineering and Technology, Hanoi, 10000, Vietnam \\
\email{nguyenhaitoan@vnu.edu.vn, datntien@vnu.edu.vn, hanv@vnu.edu.vn}\\}
\maketitle              
\begin{abstract}

Retrieval-Augmented Generation (RAG) systems commonly use chunking strategies for retrieval, which enhance large language models (LLMs) by enabling them to access external knowledge, ensuring that the retrieved information is up-to-date and domain-specific. However, traditional methods often fail to create chunks that capture sufficient semantic meaning, as they do not account for the underlying textual structure. This paper proposes a novel framework that enhances RAG by integrating hierarchical text segmentation and clustering to generate more meaningful and semantically coherent chunks. During inference, the framework retrieves information by leveraging both segment-level and cluster-level vector representations, thereby increasing the likelihood of retrieving more precise and contextually relevant information. Evaluations on the NarrativeQA, QuALITY, and QASPER datasets indicate that the proposed method achieved improved results compared to traditional chunking techniques.


\keywords{Retrieval Augmented Generation  \and Semantic Chunking \and Text Segmentation.}
\end{abstract}



\section{Introduction}

In the field of artificial intelligence (AI), processing unstructured data has become essential. Large Language Models (LLMs), such as OpenAI's GPT\footnote{https://openai.com/}, 
is capable of performing complex tasks and supporting a wide range of applications~\cite{ref_llmsurvey,ref_openai}. While these models are effective at generating responses and automating tasks, their performance is often limited by the quality of the data they process.

Updating these models via fine-tuning or other modifications can be challenging, particularly when dealing with large text corpora~\cite{ref_knowgraphs2021,ref_rag2021}. One common approach to address this issue involves dividing large volumes of text into smaller, manageable chunks. This method is widely used in question-answering systems, where splitting texts into smaller units improves retrieval accuracy~\cite{ref_wikiqa}. This retrieval-based method, known as Retrieval-Augmented Generation (RAG)~\cite{ref_rag2021}, enhances LLMs by allowing them to reference external knowledge, making it easier to ensure that the retrieved information is up-to-date and domain-specific. However, current retrieval-augmented methods have limitations, particularly in the chunking process. Traditional chunking approaches often fail to capture sufficient semantic meaning as they do not account for the underlying textual structure. This limitation becomes especially problematic for answering complex queries that require understanding multiple parts of a document, such as books in NarrativeQA~\cite{ref_narrativeqa2017} or research papers in QASPER~\cite{ref_qasper2021}.

To address these challenges, we propose a novel framework that improves the retrieval process by generating chunks that either capture local context (segments) or clusters of related segments that represent higher-level semantic coherence. The key components of this framework are:

\begin{enumerate}
    \item \textbf{Text Segmentation}: A supervised text segmentation model is applied to divide the document into smaller, coherent segments. This ensures that each segment preserves meaningful local context and is not cut off inappropriately.
    
    \item \textbf{Chunk Clustering}: After segmentation, unsupervised clustering combines related segments based on semantic similarity and their relative positions. This process creates clusters that maintain sequential structure and capture broader semantic relationships between segments.

    \item \textbf{Multiple-Vector Based Retrieval}: Each text chunk is represented by multiple vectors: several for individual segments within the chunk, and one for the cluster itself. This approach provides more options for matching during retrieval, as having multiple vectors to compare against increases the likelihood of finding a more precise match, whether based on specific segment details or broader cluster context.
\end{enumerate}

\section{Related Work}

RAG enables LLMs to access external knowledge during inference, improving their performance on domain-specific tasks. Research by Gao et al. \cite{ref_ragsurvey2023} demonstrates that RAG enhances various Natural Language Processing (NLP) tasks, particularly in improving factual accuracy. Consequently, several studies have focused on to enhance RAG's performance, either by refining the prompts used for generation \cite{ref_promptsurvey2023} or by incorporating diverse retrieval strategies~\cite{ref_hypa2024}.

A key factor in optimizing RAG systems is how documents are chunked for retrieval. Various traditional chunking methods have been developed, with open-source frameworks like LangChain\footnote{https://www.langchain.com/} and LlamaIndex\footnote{https://docs.llamaindex.ai/en/stable/} offering techniques for splitting and filtering documents. Fixed-size chunking~\cite{ref_fixedsizechunking}, which divides text into equal-length segments, is straightforward to implement but often fails to capture the semantic or structural nuances of the text. Recursive splitting~\cite{ref_langchain_recursive}, which segments text based on markers like newlines or spaces, can be effective when documents have clear formatting. Semantic chunking~\cite{github}, which groups sentences based on cosine similarity in embedding space, produces more coherent chunks, but it can sometimes lead to inconsistent boundaries or miss broader context by focusing too narrowly on sentence-level similarity.

Recent research has made strides in improving chunk retrieval and ensuring more contextually relevant chunks in RAG systems. LongRAG~\cite{ref_longrag2024} proposes using longer retrieval units, such as entire Wikipedia documents, to reduce the overall size of the corpus. In contrast, RAPTOR~\cite{ref_raptor2024} creates multi-level chunk hierarchies, progressing from detailed content at the leaf nodes to more abstract summaries at higher levels, thereby maintaining relevance while enabling the model to handle both granular and high-level content.

Another area of improvement in RAG is the integration of Knowledge Graphs (KGs)~\cite{ref_knowgraphs2021}. KGs enhance retrieval accuracy by linking related entities and concepts. For instance, GraphRAG~\cite{ref_graphrag2024} uses entity extraction and query-focused summarization to organize chunks. However, this can disrupt the natural flow of the text by grouping chunks from different sections. In contrast, our approach prioritizes cohesive chunks that maintain the original text structure. By using unsupervised clustering based on text segmentation, our chunks preserve both semantic unity and sequential order, leading to a more coherent retrieval process.

Chunking and text segmentation share similar goals, as both methods aim to divide text into manageable, coherent units. Early text segmentation approaches, such as TextTiling~\cite{ref_texttiling1997} and TopicTiling~\cite{ref_topictiling2012}, used unsupervised methods to detect shifts in lexical cohesion. With advancements in neural networks, supervised models have emerged, such as SECTOR~\cite{ref_sector2019} and S-LSTM~\cite{ref_slstm2020}, which leverage Long Short-Term Memory (LSTM) or bidirectional LSTMs to predict segment boundaries from labeled data. However, despite these advancements, text segmentation has traditionally been treated as a standalone task. Our method integrates text segmentation to improve document chunking, leading to more accurate and coherent retrieval.

\section{Hierarchical Text Segmentation Framework for RAG}

\begin{figure}[t]
    \centering
    \includegraphics[width=\textwidth]{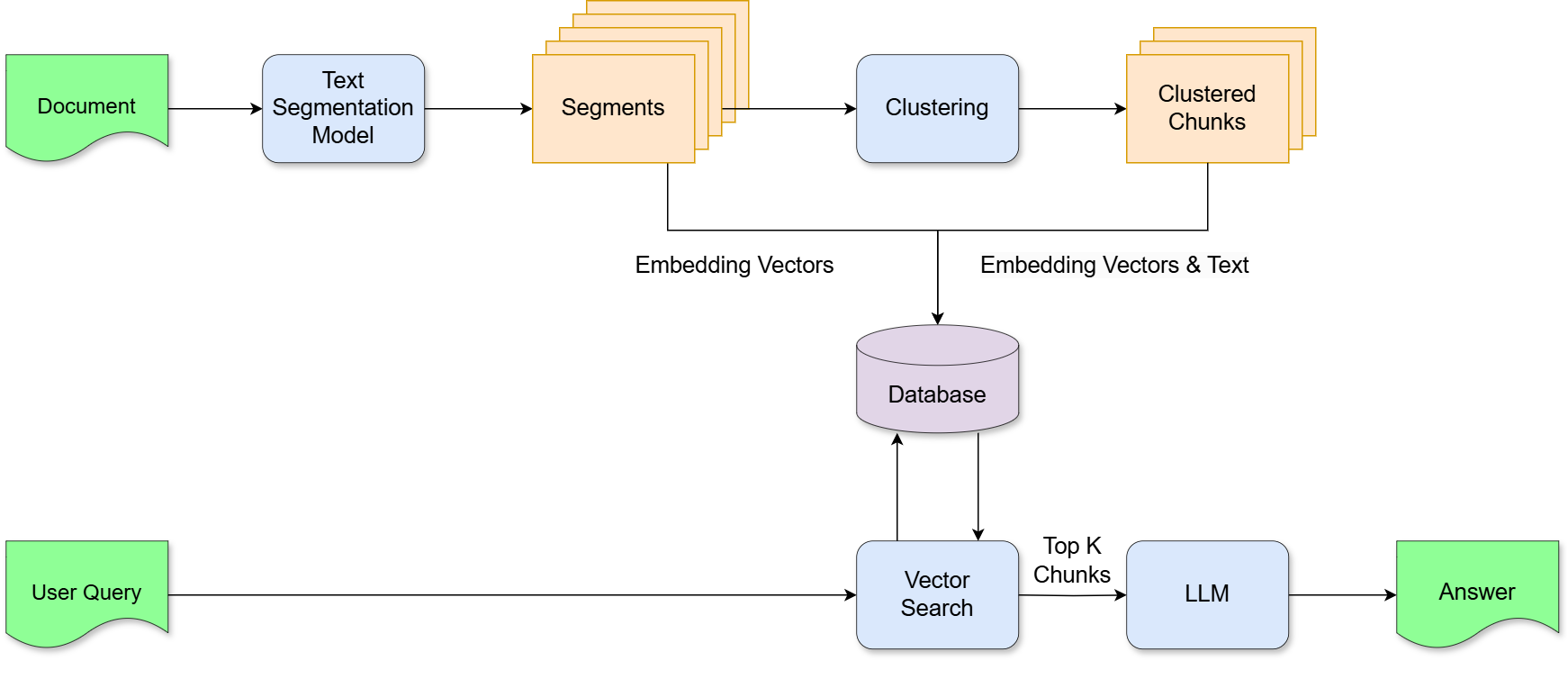}
    \vspace{0.15cm}
    \caption{Overview of our framework, incorporating text segmentation during the indexing process.}
    \label{fig:pipeline}
\end{figure}

\subsection{Overview}
Fig. \ref{fig:pipeline} illustrates our proposed RAG framework, which introduces a hierarchical segmentation and clustering pipeline to enhance chunking accuracy and retrieval relevance. During the indexing phase, each document $D$ is segmented into coherent segments $S_i$, and related segments are grouped into clusters $C_j$. Both segment and cluster embeddings are then computed and stored, as represented by the following equation:

\begin{equation}
\text{Indexing}(D) = \{(E_s, E_c) \mid E_s = f_{\text{segment}}(S_i), E_c = f_{\text{cluster}}(C_j)\}
\end{equation}

In the retrieval phase, for each chunk $C_i$, the system computes the cosine similarity between the query $q$ and all embeddings associated with the chunk, including both segment embeddings $E_s$ and cluster embeddings $E_c$. The most relevant embeddings are selected based on similarity scores, calculated as:

\begin{equation}
    \text{cos}(q, C_i) = \max(\text{cos}(q, E_{s1}), \dots, \text{cos}(q, E_{sm}), \text{cos}(q, E_c))
\end{equation}

where $E_{s1}, E_{s2}, \dots, E_{sm}$ are segment embeddings for chunk $C_i$, and $E_c$ is the cluster embedding for chunk $C_i$. The system ranks the similarity scores and selects the top-$k$ chunks for processing by the LLMs to generate the response.

\begin{figure}[b]
    \centering
    \includegraphics[width=0.75\textwidth]{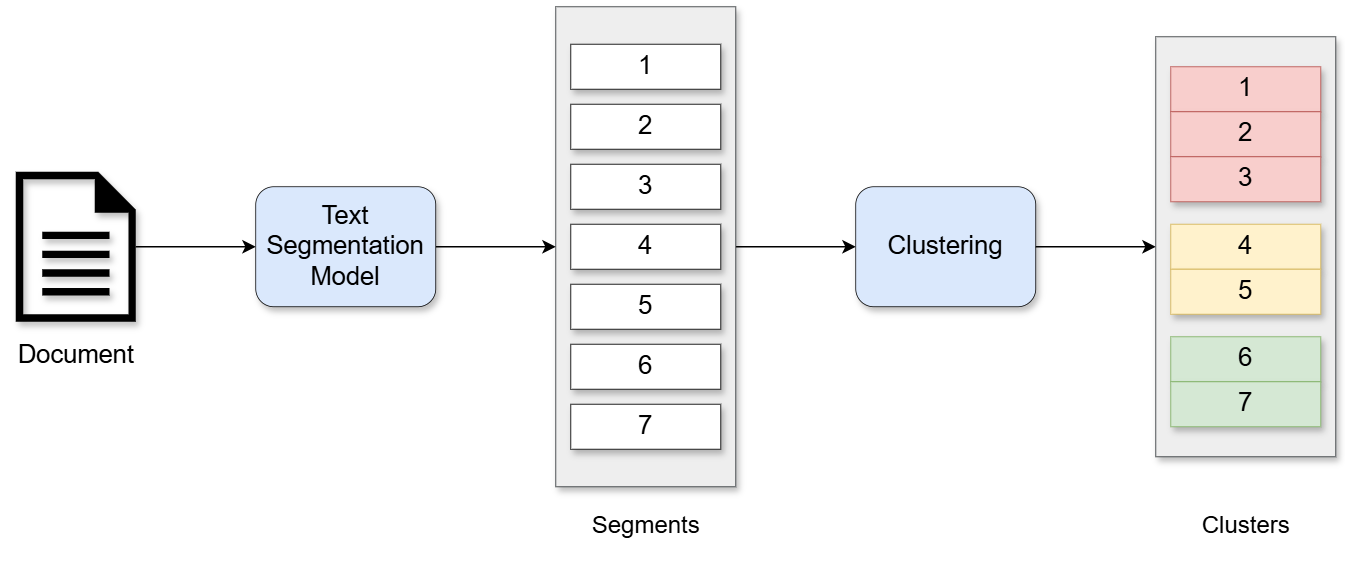}
    \caption{Illustration of the framework's chunking strategy: The text is first segmented into coherent parts using a text segmentation model. These segments are then clustered based on semantic similarities and their sequential order.}
    \label{fig:Indexing}
\end{figure}

\subsection{Text Segmentation and Clustering: A Bottom-Up Approach}
In theory, a hierarchical structure would naturally suit a top-down segmentation approach, where the document is first divided into broader sections and then broken down into smaller units. However, due to the limitations of current text segmentation models—such as the lack of multi-level training data and difficulties with processing long documents—we propose a bottom-up approach.

This bottom-up approach starts by using supervised methods to identify smaller, cohesive segments, which are then grouped into larger, meaningful units through unsupervised clustering techniques. While top-down segmentation may seem intuitive, especially for capturing hierarchical relationships, current models struggle with the complexity of longer texts like books or research articles. Fig. \ref{fig:Indexing} shows the structure of this approach.
The bottom-up approach works well with RAG’s retrieval mechanism, which does not depend on a strict sequential structure. In RAG, chunks are retrieved based on their relevance to the query, allowing more flexibility in building document representation from smaller units.

\subsubsection{Text Segmentation}
\label{Text_seg}
\begin{figure}[!b]
    \centering
    \includegraphics[width=\textwidth]{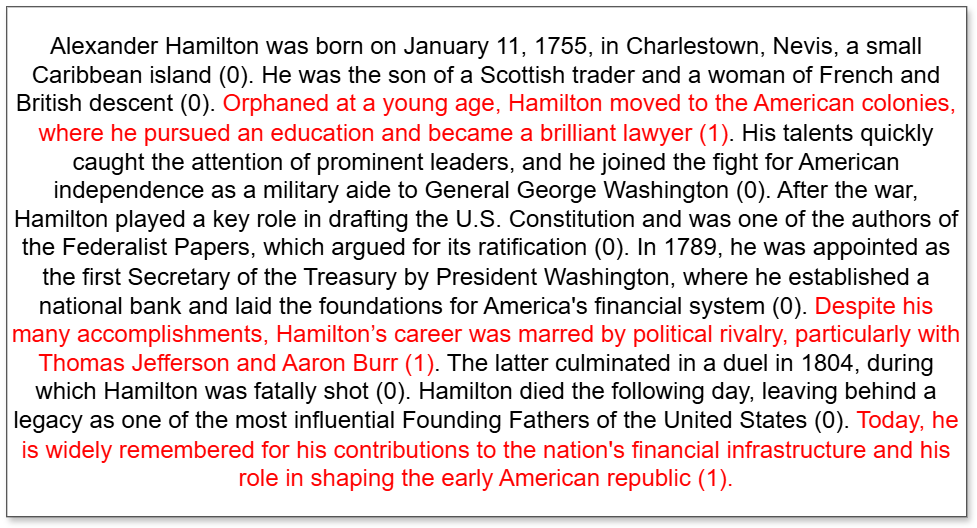}
    \caption{A biography of Alexander Hamilton is being predicted by a text segmentation model. The model predicts segment boundaries by labeling each sentence with a 1 or a 0, where 1 marks the end of a segment and 0 otherwise.}
    \label{fig:hamilton}
\end{figure}

The model used in our research, introduced by Koshorek et al. \cite{ref_textsegmentation2018}, is a neural network designed for supervised text segmentation. It predicts whether a sentence marks the end of a section by using a bidirectional LSTM to process sentence embeddings. These embeddings are generated in a previous layer, where another bidirectional LSTM processes the words in each sentence and applies max-pooling to produce fixed-length representations. The model is trained to label each sentence as either a '1' (indicating the end of a section) or a '0' (indicating continuation), by learning segmentation patterns from the training data. Fig. \ref{fig:hamilton} demonstrated an example of how a document is segmented. 

\subsubsection{Clustering}
\label{clustering}

We adapted the clustering method from Glavias et al.~\cite{ref_graphrag2024}, where instead of clustering sentences into segments, we grouped segments into cohesive clusters. The process is outlined as follows:

\begin{enumerate}
    \item \textbf{Graph Construction:} After text segmentation, each segment is represented as a node in a relatedness graph \( G = (V, E) \), where \( V \) consists of the segments. An edge is added between two segments \( S_i \) and \( S_j \) if their similarity exceeds a predefined threshold. The threshold \( \tau \) is set as:

    \begin{equation}
       \tau = \mu + k \cdot \sigma 
    \end{equation}

    where \( \mu \) is the mean similarity between all segments, \( \sigma \) is the standard deviation, and \( k \) is a parameter that controls the sensitivity of the connections between segments.

    \item \textbf{Maximal Clique Detection:} The task is to identify all maximal cliques in the graph \( G \), which are then stored in a set \( Q \).
    
    \item \textbf{Initial Clustering:} An initial set of clusters is created by merging adjacent segments that are part of at least one clique \( Q \in Q \) in graph \( G \)

    \item \textbf{Merge Clusters:} Adjacent clusters \( c_i \) and \( c_{i+1} \) are merged if there is at least one clique \( Q \in Q \) containing at least one segment from \( c_i \) and one from \( c_{i+1} \). Table~\ref{tab:merge_example} provides an illustration of this merging process.
    
    \item \textbf{Final Merging:} Any remaining single-sentence clusters are merged with the nearest neighboring cluster, based on cosine similarity, to ensure no isolated segments remain.
    
    \item \textbf{Cluster Embedding:} Once clusters are finalized, embeddings for each cluster are calculated by applying mean pooling over the vector representations of the segments within the cluster.
\end{enumerate}
\vspace{-20px}

\begin{table}[b]
\caption{Example of merging segments into clusters from cliques.}
\label{tab:merge_example}
\centering
\begin{tabular}{|c|c|}
\hline
\textbf{Step}      & \textbf{Sets}                \\ \hline
Cliques $Q$        & \{1, 2, 6\}, \{2, 4, 7\}, \{3, 4, 5\}, \{1, 6, 7\}  \\ \hline
Init. clus.         & \{1, 2\}, \{3, 4, 5\}, \{6, 7\} \\ \hline
Merge clus.         & \{1, 2, 3, 4, 5\}, \{6, 7\}  \\ \hline
\end{tabular}
\end{table}

\section{Experiments} \label{experiments}

\subsection{Datasets}

Our experiments were conducted on three datasets: NarrativeQA, QuALITY, and QASPER.

NarrativeQA contains 1,572 documents, including books and movie transcripts~\cite{ref_narrativeqa2017}. The task requires a comprehensive understanding of the entire narrative to answer questions accurately, testing the framework’s ability to comprehend and process longer, complex texts. We assess performance using ROUGE-L, BLEU-1, BLEU-4, and METEOR metrics.

The QuALITY dataset~\cite{ref_quality} consists of multiple-choice questions, each accompanied by a context passage in English. This dataset is particularly useful for testing the retrieval effectiveness of the system, as answering these questions often requires reasoning across an entire document. Accuracy is used as the evaluation metric for this dataset.

Lastly, QASPER is a benchmark dataset designed for question-answering tasks in scientific NLP papers~\cite{ref_qasper2021}, where the answers are embedded within full-text documents, rather than being located in a specific section or abstract. Performance is measured using the F1 score.

\subsection{Experimentation Settings}

For our experiments, we used the GPT-4o-mini\footnote{https://platform.openai.com/docs/models/gpt-4o-mini} model as the reader and the BAAI/bge-m3\footnote{https://huggingface.co/BAAI/bge-m3} model for embedding generation. The embeddings were stored and retrieved using FAISS\footnote{https://github.com/facebookresearch/faiss}, an efficient vector database for similarity search.

We evaluated our chunking strategies by testing different average chunk sizes: 512, 1024, and 2048 tokens. Two retrieval methods were tested: one combining segment and cluster vector search, and one using cluster vector search alone. These strategies were compared against fixed-size chunking baselines of 256, 512, 1024, and 2048 tokens, as well as the semantic chunking method by Greg Kamradt~\cite{github}, which uses an average embedding size of 256 tokens.

To maintain consistent input size for the LLM across different chunking strategies, we retrieved a proportional number of chunks based on the segment length. Specifically, we retrieved 20 chunks for 256-token chunks, 8 chunks for 512-token chunks, 4 chunks for 1024-token chunks, and 2 chunks for 2048-token chunks. This approach ensured that the total number of tokens retrieved approximately 4096, allowing for a fair comparison across all chunking strategies.

\subsection{Chunking Setup}

\textbf{Text Segmentation} As mentioned earlier in section~\ref{Text_seg}, the segmentation model we are using operates on two levels. The sentence embedding level is a two-layer bidirectional LSTM with an input size of 300 and a hidden size of 256, generating sentence representations through max-pooling over the outputs. The classifier level is similar to the first, but with double the input size (512), classifying segment boundaries based on labeled training data.

The model was trained using stochastic gradient descent (SGD) with a batch size of 32 over 20 epochs, using 100,000 documents from the Wiki727k dataset\\~\cite{ref_textsegmentation2018}. We optimized for cross-entropy loss, classifying whether a sentence marks the end of a segment. Early stopping was applied after 14 epochs, based on the validation loss plateauing.

In evaluating the text segmentation model, we used the $p_k$ metric as defined by Beeferman et al. \cite{ref_pk}, which measures the probability of error when predicting segment boundaries at random points in the text. Simply put, a lower $p_k$ score means the model is better at accurately identifying where segments end. The model achieved a $p_k$ score of 35 on the WIKI-50 test set~\cite{ref_textsegmentation2018}, compared to the original paper’s score of 20. However, it’s important to note that our focus here is not primarily on optimizing text segmentation but on enhancing retrieval through segmentation-chunking integration. As a result, we used a smaller dataset and fewer training epochs, while the original paper used much larger data and more intensive training. This trade-off allowed us to focus on the RAG framework while maintaining reasonable segmentation accuracy.

\textbf{Clustering} We set the \( k \)-values to control the number of clusters, as outlined in Section~\ref{clustering}, ensuring alignment with the average chunk size. For an average of 512 tokens, we used \( k=1.2 \), for 1024 tokens, \( k=0.7 \), and for 2048 tokens, \( k=0.4 \). Lower \( k \)-values directly reduce the number of clusters, resulting in larger token sizes per cluster.

\subsection{Retrieval Results}

As shown in Table~\ref{tab:qa_qu_metrics_result} and Table~\ref{tab:nq_metrics_result}, our segmentation-clustering method performs better than the other chunking strategies across all three datasets. For NarrativeQA, the average 1024-token segment-cluster method achieves the highest ROUGE-L score of 26.54, outperforming both the base and semantic methods. Additionally, it shows an improvement in METEOR score, reaching 30.26. In the QASPER dataset, the 1024-token segment-cluster method again yields the best results, with an F1 score of 24.67. Similarly, in the QuALITY dataset, the segment-cluster method with an average of 512 tokens attains the highest accuracy of 63.77, outperforming the base 512-token method's accuracy of 60.23.

While larger chunk sizes, such as the 2048-token configuration, might intuitively seem to provide more context, the results show diminishing returns in performance. This drop in scores is likely due to the increased size of each chunk, which makes the chunks more difficult to process and dilutes their coherence. As chunks grow larger, they capture more information but can become too broad, causing the reader model to lose focus on query-relevant details.

\begin{table}[!b]
\centering
\vspace{10px}
\caption{\textbf{Performance on QASPER and QuALITY}: Evaluation of various chunking strategies on the QASPER and QuALITY datasets. The segmentation-clustering approach yields the highest F1 score on QASPER with 1024-token average segmentation and clustering, and the highest accuracy on QuALITY with 512-token average segmentation and clustering.}
\label{tab:qa_qu_metrics_result}
\setlength{\tabcolsep}{5pt} 
\resizebox{\textwidth}{!}{%
\begin{tabular}{@{}ccccc@{}}
\toprule
\textbf{Chunk size} & \textbf{Top-k Chunks} & \textbf{Methods} & \textbf{F1 (QASPER)} & \textbf{Accuracy (QuALITY)} \\ \midrule
\multirow{2}{*}{256}  & \multirow{2}{*}{20}         & Base   & 19.28    & 58.16       \\
                                  &                 & Semantic & 18.07    & 57.23       \\ \midrule
\multirow{3}{*}{512} & \multirow{3}{*}{8}         & Base   & 20.33    & 60.23     \\
                                  &                 & Cluster Only & 21.64   & 62.36       \\
                                  &                 & Segment + Cluster  & 21.95   & \textbf{63.77}       \\ \midrule
\multirow{3}{*}{1024} & \multirow{3}{*}{4}  & Base   & 22.07    & 58.23      \\
                                  &                 & Cluster Only  & 23.31   & 58.84      \\
                                  &                 & Segment + Cluster   & \textbf{24.67}   &    59.08 \\ \midrule
\multirow{3}{*}{2048} & \multirow{3}{*}{2}  & Base   & 22.05    & 57.54      \\
                                  &                 & Cluster Only  & 22.76   & 57.71     \\
                                  &                 & Segment + Cluster   & 23.89   &    58.85 \\ \bottomrule
\end{tabular}%
}

\end{table}

\begin{table}[t]
\centering
\caption{\textbf{Performance on NarrativeQA}: Comparison of various chunking strategies on the NarrativeQA dataset. The 1024-token average segmentation and clustering outperforms all other chunking strategies across all metrics.}
\label{tab:nq_metrics_result}
\setlength{\tabcolsep}{5pt} 
\resizebox{\textwidth}{!}{%
\begin{tabular}{@{}ccccccc@{}}
\toprule
\textbf{Chunk size} & \textbf{Top-k Chunks} & \textbf{Methods} & \textbf{ROUGE-L} & \textbf{BLEU-1} & \textbf{BLEU-4} & \textbf{METEOR} \\ \midrule
\multirow{2}{*}{256}  & \multirow{2}{*}{20}          & Base   & 22.21    & 16.99      & 5.06      & 27.11 \\
                                   &                    & Semantic & 22.5    & 16.55      & 5.51      & 26.56 \\ \midrule
\multirow{3}{*}{512} & \multirow{3}{*}{8}         & Base   & 23.16    & 17.17      & 5.77      & 27.13 \\
                                   &                    & Cluster Only   & 24.12    & 17.91      & 6.55      & 27.56 \\
                                   &                    & Segment + Cluster  & 24.67   & 18.97   
                                   & 6.83      & 28.64 \\ \midrule
\multirow{3}{*}{1024} & \multirow{3}{*}{4}  & Base   & 23.86    & 18.05      & 6.59     & 27.12 \\
&                    & Cluster Only & 25.15   & 19.28   
                                   & 6.97      & 29.05 \\
                                   &                    & Segment + Cluster   &  \textbf{26.54}    & \textbf{20.03}      & \textbf{7.58}      & \textbf{30.26} \\ \midrule

\multirow{3}{*}{2048} & \multirow{3}{*}{2}  & Base   & 23.53    & 17.65      & 6.29    & 27.02 \\
&                    & Cluster Only  & 25.67   & 19.13   
                                   & 6.80      & 29.64 \\
                                   &                    & Segment + Cluster   & 26.39    & 19.62      & 7.38    & 30.07 \\ \bottomrule
\end{tabular}%
}
\end{table}

\begin{figure}[!b]
  \centering %
\begin{mybox}[The Olympic Gene Pool]
\begin{obeylines}
\textbf{Question: }The author believes that athletic ability changes over time mainly due to?

1. Top athletes having fewer children
2. Innate factors
3. \textbf{Environment}
4. Natural selection and genetics
\DrawLine
\textbf{512 Segment + Cluster:} It is scarcely surprising that Ethiopian or Kenyan distance runners do better than everyone else .... Environmental differences between the two groups could account for differing levels of athletic success ... Better health care and practicing condition affects athletic ability directly 
\textbf{Answer: 3}

\DrawLine 
\textbf{Base 512:} We know that the inheritance of extra fingers or toes is determined genetically ... Perhaps way, way back in human history, when our forebears were still fleeing saber-toothed tigers, natural selection for athletic prowess came into play... Indeed, the laws of natural selection probably work against athletes these days.
\textbf{Answer: 2}
\vspace{6pt}
\end{obeylines}
\end{mybox}
\vspace{-5px}
\caption{Retrieved chunks based on multiple chunking strategy for the question about the story The Olympic Gene Pool.}
\label{fig:prompt_hypo}
\end{figure}
We believed that using cohesive segments significantly improves the retrieval process by ensuring that meaningful units of text are retrieved. Traditional chunking often retrieves fragmented ideas due to arbitrary chunking, resulting in disjointed answers. Our Segment-Cluster method addresses this issue by grouping related segments, even if they are not adjacent. Fig.~\ref{fig:prompt_hypo} shown that this approach captures broader themes, such as training and healthcare factors. While these factors may not seem relevant to the query at first glance, they together provide better context. As a result, our method retrieves more coherent and contextually relevant information, leading to improved accuracy and overall answer quality.


\section{Conclusion}




This paper introduces a framework that integrates hierarchical text segmentation with retrieval-augmented generation (RAG) to improve the coherence and relevance of retrieved information. By combining segmentation and clustering in chunking, our method ensures that each chunk is semantically and contextually cohesive, addressing the limitations of traditional, fixed-length chunking.

Experiments showed our approach enhances retrieval accuracy and answering performance compared to traditional chunking. While a top-down segmentation approach could be ideal, current model limitations favor a bottom-up combination of supervised and unsupervised techniques. Future work may explore multi-level segmentation to streamline hierarchical representation in RAG or use enriched segments in knowledge graph construction to improve entity relationships and clustering accuracy.

\appendix


\end{document}